\documentclass[10pt,twocolumn,letterpaper]{article}

\usepackage{cvpr}              

\usepackage{colortbl}
\usepackage{multirow}
\usepackage{makecell}
\PassOptionsToPackage{prologue,dvipsnames}{xcolor}

\newcommand{\red}[1]{{\color{red}#1}}

\usepackage{pifont}
\usepackage{amsmath}
\usepackage{bm}
\newcommand{\xmark}{\text{\ding{55}}}  
\definecolor{darkgreen}{RGB}{0,127,0}
\definecolor{darkred}{RGB}{200,0,0}
\def\GC{\textcolor{darkgreen}{\checkmark}}
\def\RX{\textcolor{darkred}{\xmark}}
\newcommand{\btn}[1]{{\colorbox{green!15}{\textbf{#1}}}}
\newcommand{\stn}[1]{{\colorbox{yellow!15}{\textbf{#1}}}}
\newcommand{\rbt}[1]{{\red{\scriptsize\textbf{#1}}}}
\newcommand{\gat}[1]{{\textcolor{darkgreen}{$\downarrow$\scriptsize\textbf{#1}}}}

\definecolor{cvprblue}{rgb}{0.21,0.49,0.74}
\usepackage[pagebackref,breaklinks,colorlinks,allcolors=cvprblue]{hyperref}

\def\paperID{3349} 
\def\confName{CVPR}
\def\confYear{2026}

\title{Pip-Stereo: Progressive Iterations Pruner for Iterative Optimization based Stereo Matching}

\author{Jintu Zheng\footnotemark[1]~~\footnotemark[2], ~~Qizhe Liu\footnotemark[2], ~~HuangXin Xu, ~~Zhuojie Chen\\
[2mm]
\includegraphics[scale=0.03]{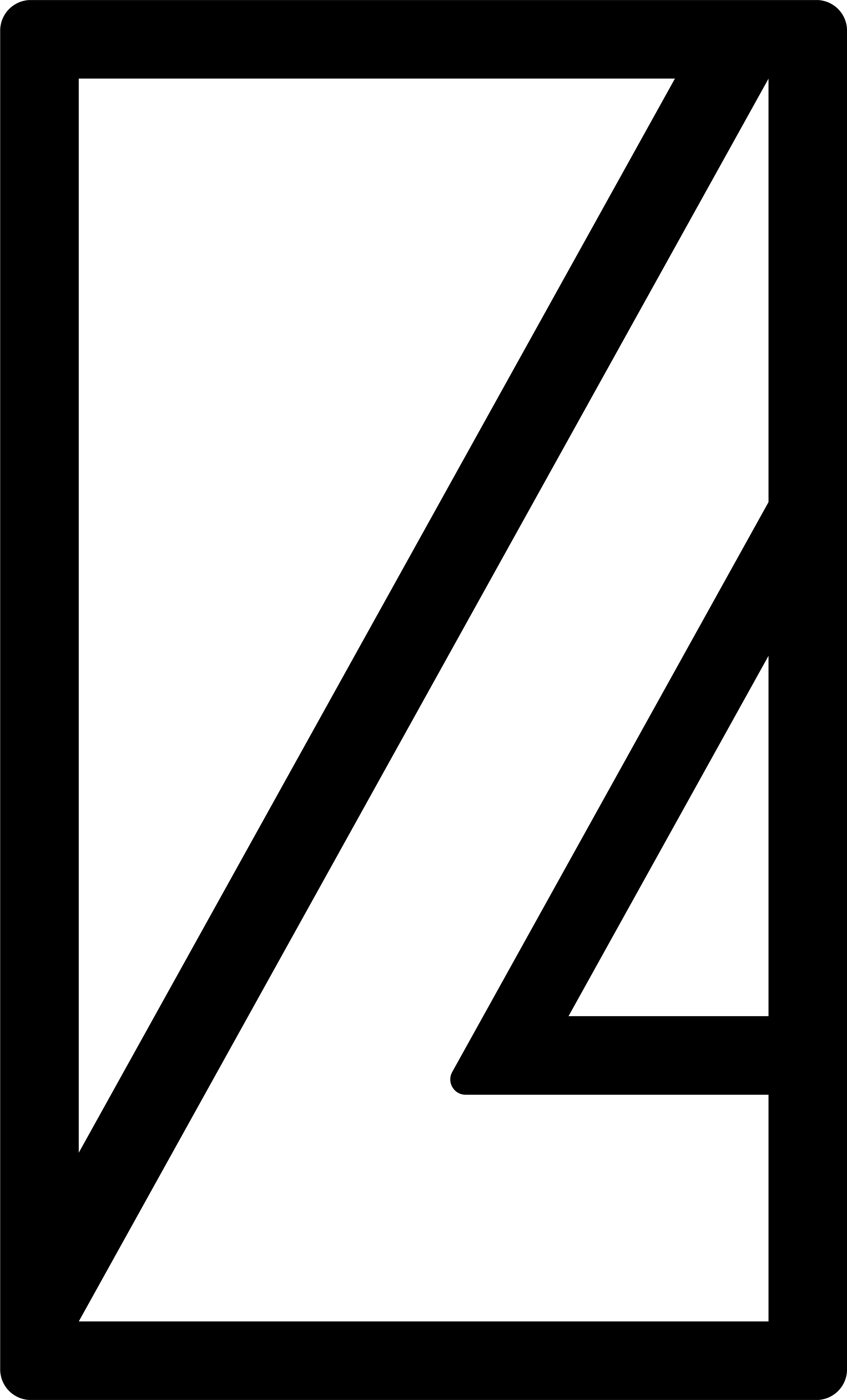}
\\
ARIDGE XPENG
\\
{\tt\small zhengjt@aridge.com}
}

\begin{document}
\maketitle

\renewcommand{\thefootnote}{\fnsymbol{footnote}}
\footnotetext[1]{Corresponding author and team leader.}
\footnotetext[2]{Both work on core development, as co-first authors.}

\begin{abstract}
While iterative stereo matching achieves high accuracy, its dependence on Recurrent Neural Networks (RNN) hinders edge deployment, a challenge underexplored in existing researches. We analyze iterative refinement and reveal that disparity updates are spatially sparse and temporally redundant. First, we introduce a progressive iteration pruning strategy that suppresses redundant update steps, effectively collapsing the recursive computation into a near-single-pass inference. Second, we propose a collaborative monocular prior transfer framework that implicitly embeds depth priors without requiring a dedicated monocular encoder, thereby eliminating its associated computational burden. Third, we develop FlashGRU, a hardware-aware RNN operator leveraging structured sparsity and I/O-conscious design, achieving a 7.28$\times$ speedup, 76.6\% memory peak reduction and 80.9\% global memory requests reduction over natvie ConvGRUs under 2K resolution. Our PipStereo enables real-time, high-fidelity stereo matching on edge hardware: it processes 320$\times$640 frames in just 75ms on an NVIDIA Jetson Orin NX (FP16) and 19ms on RTX 4090, matching the accuracy of large iterative based models, and our generalization ability and accuracy far exceeds that of existing real-time methods. Our embedded AI projects will be updated at: \textcolor{magenta}{https://github.com/XPENG-Aridge-AI}.
\end{abstract}    
\section{Introduction}
\label{sec:intro}
\begin{figure*}[t]
    \centering
    \includegraphics[width=1.0\textwidth]{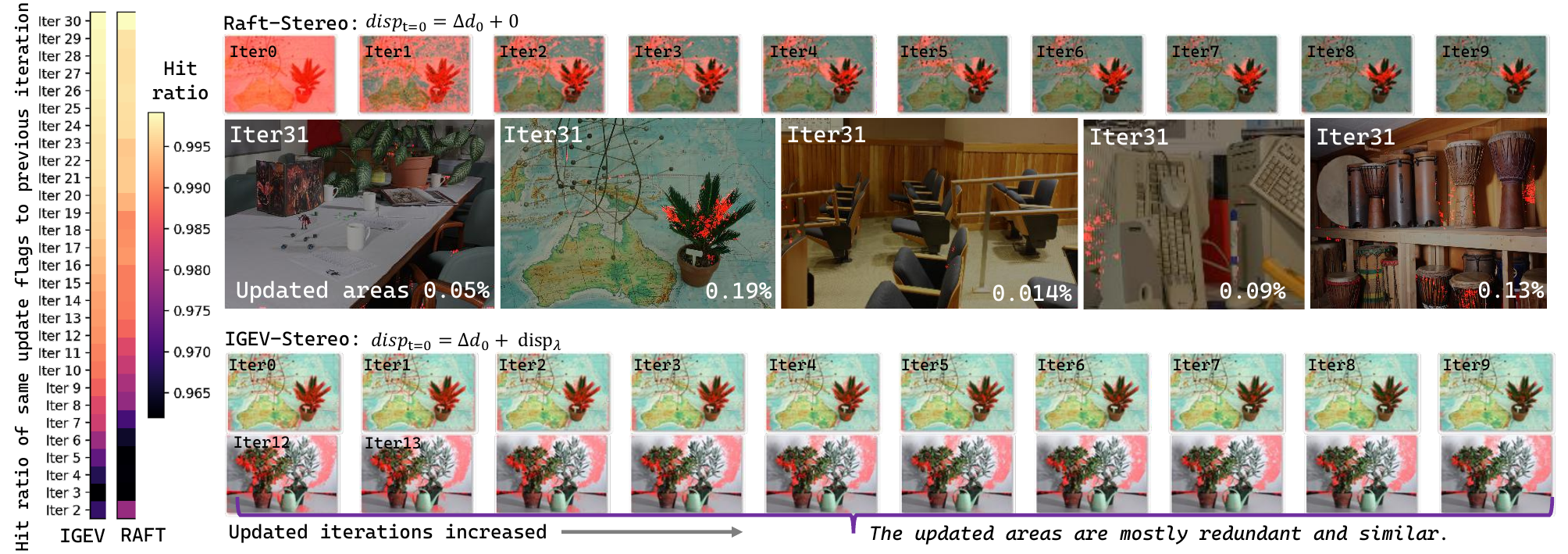}
    \caption{\textbf{Iteration update positions and hit ratios on Middlebury \citep{middlebury} test set.} Red pixels denote updates; left color bars show consistency with the previous iteration. Refinements are sparse, highly redundant, and affect only a small fraction of the image.}
    \label{fig:motivation}
    \vspace{-10pt}
\end{figure*}

Stereo matching, a cornerstone of computational vision since the dawn of binocular perception, remains pivotal in reconstructing the three dimensional fabric of the world from dual perspectives, its enduring relevance echoing through decades of practical robotic or autonomous driving deployment.

Recently, iterative optimization-based stereo matching models, e.g., MonSter \citep{cheng2025monster,mons++}, DEFOM-Stereo \citep{jiang2025defom}, IGEV \citep{xu2025igev++,xu2023iterative}, CREStereo \citep{li2022practical} and Raft-Stereo \citep{lipson2021raft}, have consistently claimed prominent positions across a multitude of benchmarks, their ascendancy owed in no small part to a potent mechanism of Gate Recurrent Units (GRU), one kind of Recurrent Neural Network (RNN). Yet, in the crucible of real-time deployment, such recurrent paradigms are often eschewed, however alluring their accuracy may be.

Drawing upon aggregated practical experience and evaluation from our large AI-Infra teams, we identify two principal factors.
First, static graphs with iterative loops suffer from complex control flow that hinders operator fusion makes RNNs highly sensitive to quantization noise.
Second, and most critically in the context of high resolution stereo vision, RNNs impose substantially heightened memory bandwidth demands. This memory-intensive behavior becomes especially pronounced at scale, where the resolution extent of input images amplifies data movement overhead and bottlenecks. In summary, the incorporation of RNN layers significantly hinders the 
edge device deployment, an obstacle that is both profound and insidious. Crucially, this challenge cannot be captured by simplistic, scalar metrics such as model parameter count or FLOPs, which fail to reflect the real-world complexities of inference latency, memory access patterns, and hardware compatibility introduced by recurrent structures. The prevailing approaches have disproportionately centered on eliminating cost volume computation, while largely neglecting the deployment bottlenecks inherently entailed by RNN layers.

Most real-time approaches circumvent iterative refinement by streamlining costly components, most notably through innovations in feature extraction \citep{xu2020aanet,xu2021bgnet,xu2023Fast-ACVNet} and regularization architectures tailored to the cost volume \citep{tankovich2021hitnet,guo2025lightstereo,bangunharcana2021CoEx,li2024iinet}. Yet such real-time methods, relying as they do on module redesign, often exhibit a lamentable degradation in accuracy when finetuned for subdomains, falling short of the performance witnessed on public benchmarks. From the experience of engineering, we observe that non-iterative real-time approaches exhibit markedly inferior generalization and robustness compared to their iterative counterparts, even when both are engineered under comparable latency constraints. This compels us to ponder: is iterative refinement truly an indispensable instrument for attaining precision in both the training and inference of stereo matching?

We conduct an analysis of how iterative refinement influence disparity estimation to further explore. As illustrated in \cref{fig:motivation}, we select two representative paradigms: Raft-Stereo \citep{lipson2021raft}, which initiates refinement from a null disparity hypothesis, and IGEV \cite{xu2023iterative}, which bootstraps its updates from a coarse initial disparity. We analyze the spatial footprint of iterative updates on the Middlebury \citep{scharstein2014middlebury} test set, and quantifying their update flags hit ratio (i.e., 0 is no update, 1 is updated, and obtain the overlap of two flag maps). Both iterative paradigms converge on a strikingly consistent observation: refinement activity is highly sparse and overwhelmingly redundant. As shown in \cref{fig:motivation}, the hit ratio exhibits a clear upward trajectory across iterations. In the case of IGEV \cite{xu2023iterative}, it stabilizes remarkably early, surpassing and sustaining a level above 0.99 from iteration 10 onward. RAFT \citep{lipson2021raft} begins its refinement from a null disparity hypothesis, resulting in a more gradual ascent; nevertheless, by around iteration 15, its hit ratio likewise converges to a near-constant plateau exceeding 0.99. This convergence underscores a characteristic: after a brief transient phase, successive updates overwhelmingly revisit the same sparse set of pixels, signaling the onset of diminishing returns in the refinement process. By the time the 32 iterations are reached, the set of updated pixels dwindles to less than 1\% of the entire image, revealing a process in which most iterations merely echo what has already been settled.

Inspired by the aforementioned insights, we introduce three novel designs that thoughtfully reconcile the high accuracy and strong generalization inherent to iterative methods with the stringent computational and memory-access constraints imposed by edge deployment.
First, we propose a novel pruning algorithm for iterative optimization-based stereo matching. We apply successive halving to prune the total number of iterations and perform progressive feature alignment, aiming to gradually drive the recursive computation graph toward convergence within a single iteration. Under this pruning scheme, existing iterative optimization-based stereo matching methods can eliminate the I/O and computational bottlenecks of RNNs, achieving significantly accelerated inference on edge devices while preserving high accuracy, leaving all existing real-time based methods far behind.
Recent studies have compellingly demonstrated that incorporating monocular depth priors significantly enhances the capacity of stereo matching models to disambiguate ill-posed regions \citep{cheng2025monster,jiang2025defom,mons++}. However, they invariably embed a dedicated depth-prediction foundation model (as an independent feature encoder) into the overall pipeline, thereby imposing a substantial computational burden that impedes deployment on edge devices. To this end, we propose a collaborative learning paradigm that enables knowledge transfer of monocular depth priors through re-parameterized encoders block within the student branch. This design effectively offloads the computational overhead of an explicit monocular depth encoder, allowing the stereo and monocular streams to share a unified, lightweight feature extractor.
Moreover, we develop a sparse GRU operator, FlashGRU, designed with a hardware-aware philosophy that diverges fundamentally from prior real-time approaches \citep{li2024iinet,xu2020aanet,tankovich2021hitnet,guo2025lightstereo}. Our design is crafted in accordance with the inherent characteristics of contemporary GPUs and NPUs. By making the RNN parts I/O aware and leveraging structured sparsity, FlashGRU attains a threefold acceleration compared to the native Selective GRU implementation, without significantly sacrificing precision.

We validate our three designs through extensive experiments, achieving the best accuracy and speed trade-off for iterative stereo matching. Rigorous edge-hardware benchmarking and performance counter analysis reveal the memory-access bottleneck of high-resolution iterative methods, and confirm our designs inherently reduce both compute and memory demands, yielding a hardware-aware, deployment-friendly architecture.

Our contributions can be summarized as follows:
\begin{itemize}
\item We propose a novel \textbf{P}rogressive \textbf{I}terations \textbf{P}runing algorithm (\textbf{Pip-Stereo}), for iterative stereo matching methods, to eliminate the deployment bottleneck of RNN.
\item We propose a collaborative learning paradigm for monocular depth prior transfer instead of embedding a foundational depth network into inference.
\item We introduce an efficient GRU operator via structured sparsity and I/O aware (FlashGRU), which has advantages in overcome memory access wall at high resolution images.
\end{itemize}

\section{Related Work}
\noindent
\textbf{Iterative Optimization in Stereo Matching.}
Early stereo matching networks predominantly relied on cost-volume filtering \citep{cheng2022region,cheng2024coatrsnet,gu2020cascade,guo2019group,mayer2016large,shen2021cfnet,xu2023unifying} to recover disparity maps. In contrast, a new generation of methods grounded in iterative optimization \citep{lipson2021raft,xu2023iterative,li2022practical,wang2024selective,cheng2025monster,jiang2025defom} has recently demonstrated remarkable gains in accuracy. These approaches harness recurrent architectures, most notably ConvGRU layers, to progressively refine disparity estimates through multiple update steps. Broadly speaking, they fall into two dominant paradigms: the zero-initialized iterative scheme exemplified by RAFT-Stereo \citep{lipson2021raft}, which begins refinement from a null disparity hypothesis, and the regularized initialization strategy pioneered by IGEV \citep{xu2023iterative}, which bootstraps optimization from a coarse but structured initial estimate. Building upon these foundations, several powerful variants have emerged. CREStereo \citep{li2022practical} introduces a coarse-to-fine disparity propagation mechanism that aligns refinement with multi-scale geometry, while Selective-IGEV \citep{wang2024selective} incorporates importance-aware attention maps to dynamically allocate computational focus toward ambiguous or textureless regions. More recently, spurred by the rapid advances in monocular depth foundation models, works such as MonSter \citep{cheng2025monster} and DEFOM-Stereo \citep{jiang2025defom} have leveraged monocular depth priors to accelerate convergence in stereo matching—effectively injecting global geometric inductive bias into the iterative process.
Nevertheless, these sophisticated designs ill-suited for deployment on edge platforms. The latency they incur, e.g., MonSter \citep{cheng2025monster} requiring approximately 7.6 seconds per $384\times 1344$ frame on an NVIDIA Orin NX, is prohibitive for real-time applications such as autonomous driving or robotic navigation, where tens of millisecond inference is often mandatory. 
The deployment bottleneck of iterative methods mainly arises from two issues: the high cost of building and processing cost volumes, and the poor hardware efficiency of recurrent modules. While cost-volume redundancy has been extensively addressed in prior work, and lies beyond our scope, efficient recurrence remains a critical yet overlooked challenge.

\noindent
\textbf{Efficiency and Accuracy Trade-offs.}
Current approaches \citep{bangunharcana2021CoEx,guo2025lightstereo,bangunharcana2021CoEx,tankovich2021hitnet,xu2020aanet,li2024iinet,xu2021bgnet,xu2023Fast-ACVNet} that prioritize efficiency typically dispense entirely with the RNN part, substituting it with bespoke architectural innovations or alternative regression paradigms in an attempt to offset the consequent loss in accuracy. The prevailing focus of these methods lies predominantly in refining the feature extraction stage of the cost volume. For instance, LightStereo \citep{guo2025lightstereo} introduces a 2D cost aggregation mechanism; CoEx \citep{bangunharcana2021CoEx} enhances channel-wise features within the cost volume through guided excitation; HitNet \citep{tankovich2021hitnet} replaces the RNN with a multi-level cascaded propagation scheme; AANet \citep{xu2020aanet} incorporates channel attention to recalibrate cost volume representations; and IINET \citep{li2024iinet} proposes leveraging MLP to refine cost volume computation. Nevertheless, these architectures, often engineered with an overriding emphasis on structural parsimony, exhibit markedly limited adaptability when fine-tuned on data from specialized subdomains. Their robustness and generalization capabilities fall significantly short of those achieved by iterative, RNN-based formulations.
Concurrently, a parallel line of research has sought to enhance the efficiency of iterative stereo methods: exemplified by approaches such as RAFT-Stereo's \citep{lipson2021raft} slow-fast scheme, RT-MonSter \citep{mons++}, and RT-IGEV \cite{xu2025igev++}. These methods truncate the number of refinement iterations, reduce GRU levels, or downscale the backbone architecture. While such measures succeed in accelerating inference, they do so at a steep cost in accuracy. The inherent temporal dependency of recurrent mechanisms, renders them acutely sensitive to architectural simplifications.
In this paper, we address this fundamental trade-off, and propose the principled solutions that reconciles efficiency with performance.

\section{Method}
\begin{figure*}[t]
    \centering
    \includegraphics[width=1.0\textwidth]{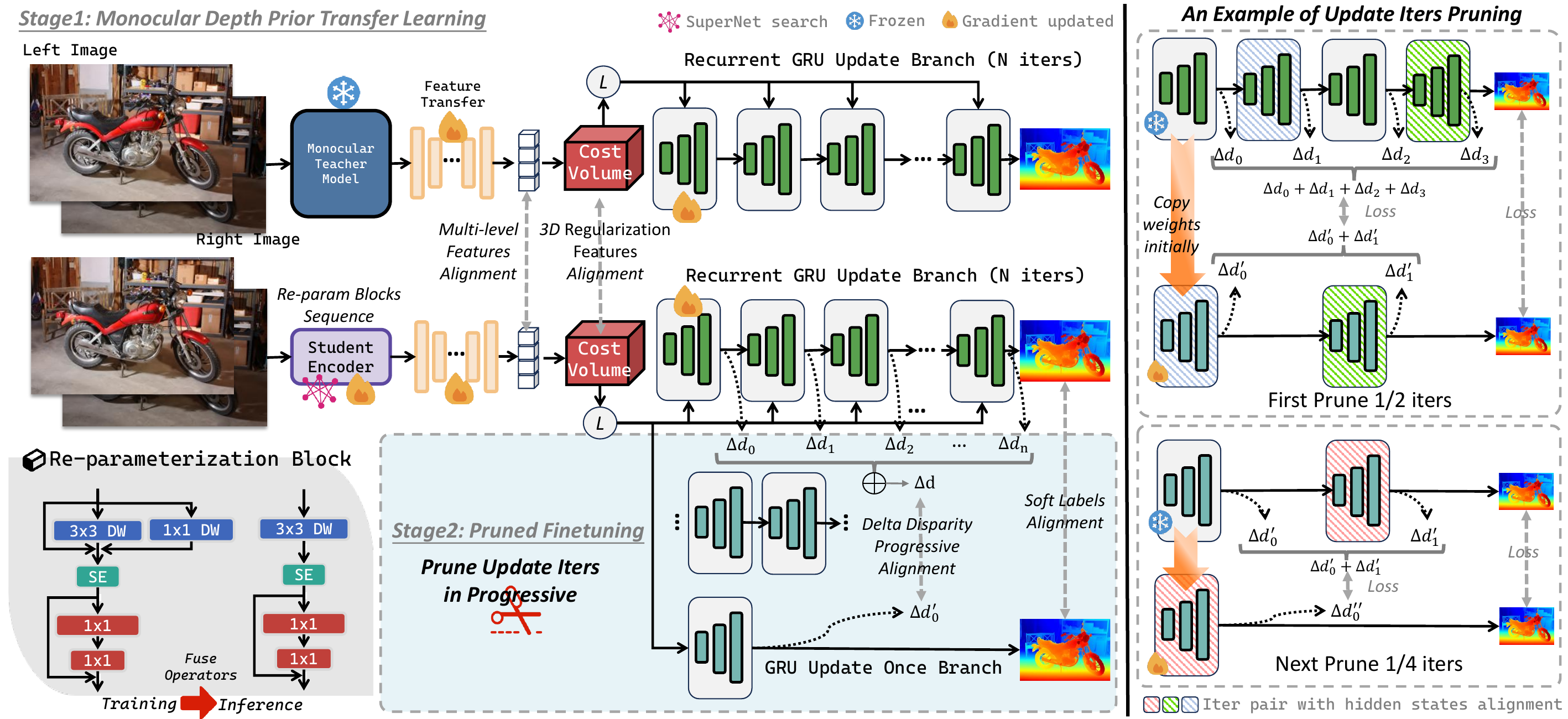}
    \caption{\textbf{Overview of two-stage training:} (1) depth prior transfer via multi-level and cost volume feature alignment, and (2) pruned finetuning with progressively fewer iterations, training only the ConvGRU modules.}
    \label{fig:pipeline}
    \vspace{-10pt}
\end{figure*}
\subsection{Overview}
Our goal is to dismantle the deployment bottleneck endemic to iterative optimization–based stereo matching methods, without compromising state-of-the-art accuracy. As depicted in \cref{fig:pipeline}, our approach departs fundamentally from conventional training paradigms through a two-stage learning strategy:
(1) Monocular depth prior transfer learning, which harnesses rich cues from pretrained monocular depth estimators to resolve ambiguities in ill-posed regions; and
(2) Pruned finetuning, which distills the iterative refinement process into a single-pass inference while preserving high-fidelity disparity predictions.
We introduce the training implementation in \cref{sec:mono}, \ref{sec:pip}, detailing each algorithm rationale and integration.
To further broaden applicability, we introduce FlashGRU, an I/O-aware, structurally sparse recurrent operator designed as a lightweight fallback for architectures resistant to single-iteration compression, such as those relying on zero-disparity initialization.

\subsection{Monocular Depth Prior Transfer Learning}
\label{sec:mono}
The objective of first stage is to transfer knowledge from a monocular depth prediction foundation model that is beneficial for stereo matching. To this end, we instantiate a teacher-student framework: the teacher network leverages a monocular foundation model (depth-anythingv2-L \cite{depthany_v2} adopted) with strong generalization capabilities.
The gradients from monocular depth teacher and student networks are co-updated in tandem. To accelerate the convergence of collaborative learning, we determine the student for adapting the teacher network through early warm-up training.

\subsubsection{Blocks Allocation via Supernet Search}
We adopt the lightweight RepViT \citep{wang2024repvit} block as the student backbone's basic unit (\cref{fig:pipeline}, bottom left). Unlike standard practice, directly finetuning a pretrained RepViT, we perform a dedicated architectural search to discover a stereo matching transfer learning specific configuration.
In iterative frameworks, the multi-scale features produced by the backbone are directly or indirectly leveraged across multiple downstream modules, for instance, fused within the context network to enrich context reasoning. We posit that the significance of features varies across resolution levels: for example, the 4$\times$ downsampled features play a pivotal role in constructing the cost volume. To this end, we perform a fine-grained supernet search to sample a student network, optimizing the capacity of allocateing the feature representation of each layer, determining precisely how much high-frequency details and how much abstract contextual knowledge should be preserved at every stage of the encoding hierarchy.

We begin by initializing the student from the M2-3 backbone of RepViT. From this seed architecture, we clone the existing blocks to serval times, spanning two structural variants, with and without channel squeeze excitation, across four distinct resolutions, thereby expanding into a supernet. This supernet is then subjected to a brief yet effective warmup phase to stabilize its initial representations. Subsequently, we deploy a genetic algorithm to conduct a structured search over layer-wise configurations. This evolutionary process is designed to uncover both redundant blocks and those critically sensitive to accuracy, ultimately yielding an optimized block arrangement tailored to each resolution (refer more details in Appendix).

\subsubsection{Teacher-to-Student Features Alignment}
To impose meaningful feature-level guidance from teacher to student, we enforce alignment of pivotal representations along the inference pipeline, specifically, the multi-resolution contextual features and the cost volume embeddings produced by the regularization network for disparity regression. This cross-architectural correspondence is supervised through the Mean Squared Error (MSE) loss.

\subsection{Progressive Iteration Pruner (PIP)}
\label{sec:pip}
We propose a simple yet effective iterative pruning algorithm that preserves accuracy while progressively reducing the number of recurrent steps.
For convenience, we refer to the RNN with \textbf{M}ore \textbf{I}terations in each pruning process as Mi-RNN and the one with \textbf{F}ewer \textbf{I}terations as Fi-RNN. As illustrated in the right panel of \cref{fig:pipeline}, we confine finetuning exclusively to the RNN module. The Fi-RNN weights are initialized from the Mi-RNN, and a halving schedule is applied to the number of unrolled iterations. This successive pruning, iteratively collapsing a multi-step inference trajectory into a more compact counterpart, guides the architecture toward a minimal yet performant recurrent depth, with less sacrificing fidelity to the original output distribution.

Formally, let the Mi-RNN define a discrete dynamical system.
\begin{equation}
    \mathbf{z}_{t+1}^{\text{Mi-RNN}} = \mathcal{F}_{\theta}(\mathbf{z}_t^{\text{Mi-RNN}}), \quad t = 0, \dots, T-1,
\end{equation}
where $ \mathbf{z}_t \in \mathbb{R}^d $ denotes the hidden state at iteration $t$, and $T$ is the total number of iterations (e.g., $T=24 $ adopted in experiments). The Mi-RNN executes only $S = T/r$ steps (with compression ratio $r=2$):
\begin{equation}
    \mathbf{z}_{s+1}^{\text{Fi-RNN}} = \mathcal{G}_{\phi}(\mathbf{z}_s^{\text{Fi-RNN}}), \quad s = 0, \dots, S-1.
\end{equation}
We enforces skip-step equivalence: the Mi-RNN's trajectory should reproduce the aggregated effect of the Fi-RNN's full trajectory over coarse time intervals.
Specifically, let $\Psi(\cdot)$ denote the output mapping (e.g., disparity update block). We define the Mi-RNN's block-aggregated output over each $r$-step window as
\begin{equation}
    \bar{\mathbf{d}}_s^{\text{Mi-RNN}} = \frac{1}{r} \sum_{i=1}^{r} \Psi\big(\mathbf{z}_{r(s-1)+i}^{\text{Mi-RNN}}\big), \quad s = 1, \dots, S,
\end{equation}
and the Fi-RNN's per-step output as $ \mathbf{d}_s^{\text{Fi-RNN}} = \Psi(\mathbf{z}_s^{\text{Fi-RNN}})$.
The core objective is to align the cumulative outputs up to each coarse step:
\begin{equation}
    \mathcal{L}_{\text{cum}} = \sum_{s=1}^{S} \left\| \sum_{k=1}^{s} \mathbf{d}_k^{\text{Fi-RNN}} - \sum_{k=1}^{s} \bar{\mathbf{d}}_k^{\text{Mi-RNN}} \right\|_2^2.
\end{equation}
This encourages the Fi-RNN to match not only instantaneous predictions but also the evolutionary trend of the Mi-RNN's refinement process. Additionally, we supervise the final disparity directly:
\begin{equation}
    \mathcal{L}_{\text{final}} = \big\| \mathbf{d}_S^{\text{Fi-RNN}} - \Psi(\mathbf{z}_T^{\text{Mi-RNN}}) \big\|_2^2,
\end{equation}
and include hidden-state matching over the coarse grid:
\begin{equation}
    \mathcal{L}_{\text{hid}} = \sum_{s=1}^{S} \big\| \mathbf{z}_s^{\text{Fi-RNN}} - \mathbf{z}_{rs}^{\text{Mi-RNN}} \big\|_2^2.
\end{equation}
The total loss is  
\begin{equation}
\mathcal{L} =  \mathcal{L}_{\text{cum}} +  \mathcal{L}_{\text{final}} + \mathcal{L}_{\text{hid}}.
\end{equation}
From a dynamical systems perspective, PIP algorithm learns a coarse-grained operator $ \mathcal{G}_\phi $ that approximates the $ r $-fold composition $ \mathcal{F}^{(r)} = \mathcal{F} \circ \cdots \circ \mathcal{F} $, while preserving the integral characteristics of the original trajectory, ensuring that the pruned model retains the Mi-RNN's refinement effect despite reduced computational steps.
In each pruning process, PIP enables a 2$\times$ reduction in iteration count with minimal accuracy drop, and can be recursively applied for further compression (e.g., $T\to T/2 \to T/4 $).

\begin{figure}[t]
    \centering
    \includegraphics[width=0.5\textwidth]{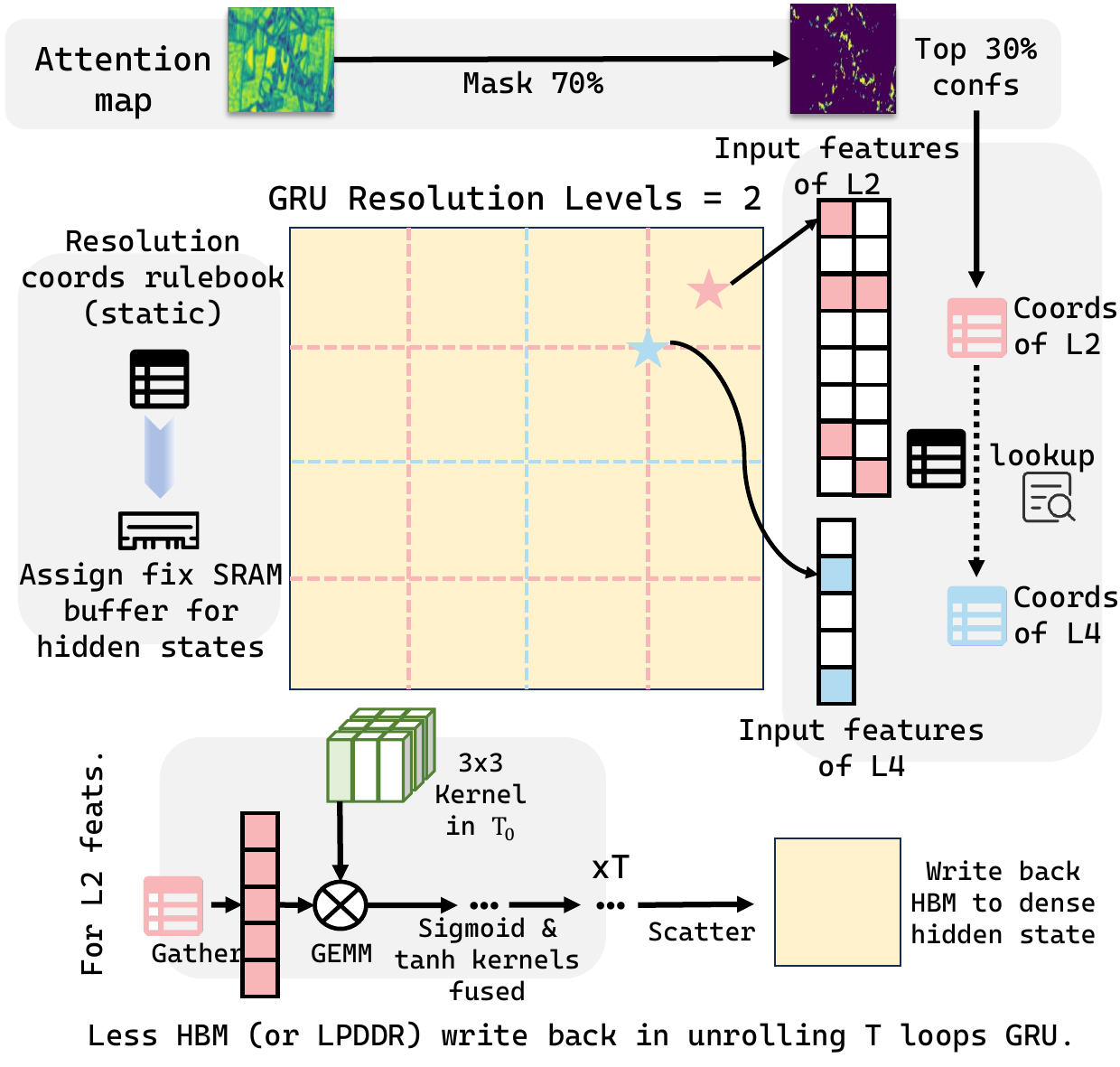}
    \caption{Illustrion of the FlashGRU in 2 resolution levels and T loops, with 70\% sparisity.}
    \label{fig:flashgru}
    \vspace{-10pt}
\end{figure}

\begin{table*}\footnotesize
\setlength{\tabcolsep}{1.5pt}
    \centering
    \begin{tabular}{l|c|c|ccc|cccc|cccc|c}
     \toprule
     \multirow{3}{*}{\textbf{Method}} 
                                      & \multirow{3}{*}{\textbf{Iters.}} 
                                      & \multicolumn{1}{c|}{\textbf{Sceneflow}\citep{sceneflow}} 
                                      & \multicolumn{3}{c|}{\textbf{ETH3D}\citep{eth3d}}  
                                      & \multicolumn{4}{c|}{\textbf{KITTI 2012}\citep{kitti2015}} 
                                      & \multicolumn{4}{c|}{\textbf{KITTI 2015}\citep{kitti2012}} 
                                      & \multirow{3}{*}{\textbf{Latency(s)}}   \\
     \cline{3-14}
&     & EPE        & Bad-1      & Bad-1      & RMSE       & Out-2      & Out-2      & Out-3      & Out-3      & D1-bg      & D1-all     & D1-bg      & D1-all     & \\
&     & All        & Noc        & All        & Noc        & Noc        & All        & Noc        & All        & Noc        & Noc        & All        &  All       & \\
     \hline
     RaftStereo\citep{lipson2021raft}                 
& 32  & 0.72       & 2.44       & 2.60       & 0.36       & 1.92       & 2.42       & 1.30       & 1.66       & 1.44       & 1.69       & 1.58       & 1.82        & 2.95\\
     IGEV\citep{xu2023iterative}                       
& 12  & 0.49       & 1.12       & 1.51       & 0.34       & 1.71       & 2.17       & 1.12       & 1.44       & 1.27       & 1.49       & 1.38       & 1.59        & 1.29\\
     SelectiveIGEV\citep{wang2024selective}             
& 12  & 0.44       & 1.23       & 1.56       & 0.29       & 1.59       & 2.05       & 1.07       & 1.38       & 1.22       & 1.44       & 1.33       & 1.55        & 1.61\\
     DEFOMStereo\citep{jiang2025defom}               
& 32  & \stn{0.42} & 0.70       & 0.78       & 0.22       & 1.43       & 1.79       & 0.94       & 1.18       & 1.25       & 1.41       & 1.15       & \btn{1.33}  & 5.05 \\
     FoundaStereo(L)\citep{foundationstereo}             
& 32  & -          & \stn{0.26} & \stn{0.48} & \stn{0.20} & -          & -          & -          & -          & -          & -          & -          & -           & 14.02 \\
     MonSter\citep{cheng2025monster}                    
& 32  & \btn{0.37} & 0.46       & 0.72       & \stn{0.20} & \stn{1.36} & \stn{1.75} & \stn{0.84} & \stn{1.09} & \stn{1.05} & \stn{1.33} & \stn{1.13} & 1.41        & 7.63\\
     MonSter++\citep{mons++}                    
& 32  & \btn{0.37} & \btn{0.25} & \btn{0.45} & \btn{0.18} & \btn{1.30} & \btn{1.70} & \btn{0.79} & \btn{1.07} & \btn{1.02} & \btn{1.29} & \btn{1.12} & \stn{1.37}  & 7.63\\
\hline
     BGNet+\citep{xu2021bgnet}                    
& \RX & 1.17       & -          & -          & -          & 2.78       & 3.35       & 1.62       & 2.03       & 1.66  & 2.01  & 1.81  & 2.19      & 0.16\\
     AANet\citep{xu2020aanet}                     
& \RX & 0.87       & 5.01       & 5.41       & 0.58       & 2.30       & 2.96       & 1.55       & 2.04       & 1.80  & 2.32  & 1.99  & 2.55      & 0.48\\
     LightStereo(S)\citep{guo2025lightstereo}              
& \RX & 0.73       & 21.82      & 22.81      & 8.15       & 3.28       & 3.93       & 1.88       & 2.34       & 1.83       & 2.11       & 2.00       & 2.30      & 0.13\\
     CoEx\citep{bangunharcana2021CoEx}                       
& \RX & 0.67       & 19.78      & 20.15      & 5.95       & 2.54       & 3.09       & 1.55       & 1.93       & 1.62       & 1.86       & 1.74       & 2.02      & 0.17\\
     FastACVNet+\citep{xu2023Fast-ACVNet}                
& \RX & 0.59       & 5.62       & 6.00       & 0.63       & 2.39       & 2.97       & 1.45       & 1.85       & 1.56       & 1.85       & 1.70       & 2.01      & 0.27\\
     HITNet(L)\citep{tankovich2021hitnet}                   
& \RX & 0.55       & 2.79       & 3.11       & 0.46       & 2.00       & 2.65       & 1.41       & 1.89       & 1.54       & 1.54       & 1.74       & 1.98      & 0.44\\
     IINet\citep{li2024iinet}                      
& \RX & 0.54       & 5.94       & 6.21       & 0.57       & 2.76       & 3.34       & 1.81       & 2.21       & 1.87       & 2.07       & 2.02       & 2.25      & 0.29\\
     RT-IGEV++\citep{xu2025igev++}
& 6   & \stn{0.52} & 3.81       & 4.45       & 0.66       & 1.93       & 2.51       & 1.29       & 1.68       & 1.34       & 1.64       & 1.48       & 1.79      & 0.39\\
     RT-MonSter++\citep{mons++}
& 4   & 0.76       & \stn{1.32} & \stn{1.46} & \stn{0.29} & \stn{1.75} & \stn{2.26} & \stn{1.07} & \stn{1.41} & \stn{1.33} & \stn{1.52} & \stn{1.47} & \stn{1.69}  & 0.79 \\
     \textbf{PipStereo (Ours)}
                & 1   
                & \btn{0.45\tiny{(-13.5\%)}} 
                & \btn{0.35\tiny{(-73.4\%)}}  & \btn{0.67\tiny{(-54.1\%)}} & \btn{0.19\tiny{(-34.4\%)}} 
                & \btn{1.60}  & \btn{1.65} & \btn{0.92} & \btn{0.94}
                & \btn{1.20}  & \btn{1.49} & \btn{1.33} & \btn{1.44}      
                & 0.44 \\
    \bottomrule
    \end{tabular}
    \vspace{-5pt}
    \caption{\textbf{In-domain comparison with PipStereo.}
Top panel lists high-accuracy iterative approaches, while bottom panel includes real-time methods.
PipStereo achieves accuracy competitive with state-of-the-art iterative methods while significantly outperforming all high-accuracy approaches in inference latency, and substantially surpassing existing real-time methods in accuracy.
The \btn{item} and \stn{item} are the best and second best in each panel. Latencies are tested in $384\times 1344$ resolution on Orin NX (25W power mode, FP32 precision).
}
\label{tab:benchmark}
\vspace{-10pt}
\end{table*}

\subsection{FlashGRU}
In paradigms with non-zero initial disparity, aggressively reducing iteration counts leads to sharp accuracy degradation. In the update block, frequent updates to the hidden state cause significant memory-bound stalls when processing high-resolution images.
We observe that these updates are temporally coherent but spatially sparse. Leveraging this property, we develop FlashGRU, a hardware I/O-aware operator that reducing expensive memory write back to compatible acceleration.

\noindent
\textbf{Multi-resolution Rulebook.}
As illustrated in \cref{fig:flashgru}, we leverage an importance map (predicted by channel-spatial attention module proposed in Selective-IGEV \cite{wang2024selective}) to select candidate update regions, enforcing sparsity constraints by applying a threshold and retaining only the top-k pixels. We observe that in ConvGRU, different update units handle distinct resolution levels, so during initialization we construct a static multi-level bidirectional index mapping table that links pixel coordinates across the multi-resolution hierarchy to enable a more compact parallel computation graph. Based on the target sparsity level, we pre-allocate contiguous GPU buffer sizes for the entire computation pipeline. This multi-resolution coordinate mapping allows us to pack sparse pixels from multiple levels as contiguously as possible into GPU buffers, thereby reducing memory fragmentation overhead and facilitating computation within a single GPU context.

\noindent
\textbf{Operator Fusion in Loops.}
We unroll the recurrent computation and implement the sequential convolutions in the GRU as a temporally fused kernel, leveraging the index mapping table to minimize the number of memory write-backs. The gain of FlashGRU would be scaled up in high-resolution inputs, and limited to the GPU on-clip memory size (or L2 cache size in edge device) in different devices.

\section{Experiment}
\subsection{Experiment Setup}
\noindent
\textbf{Declarations.}
We denote the standard version as \textbf{PipStereo}, which incorporates both Monocular Prior Transfer (MPT) and Progressively Iterative Pruning (PIP), reducing the number of refinement iterations from 32 to 1.
To assess the impact of PIP, we introduce \textbf{PipStereo-i32}, a variant that retains all 32 iterations (i.e., without PIP).
Intermediate pruning stages are denoted as \textbf{PipStereo-pk} (k=1,2,3,4), where the number of iterations is $32/2 
^k$. FlashGRU is a complementary acceleration strategy compatible with PIP. However, its speedup is marginal when only a single iteration remains; thus, we compare it against configurations with more than one iteration.
We name those with FlashGRU are \textbf{Pk-Flash}, e.g., P3-Flash retains 4 iterations.

\noindent
\textbf{Datasets.}
Follow most approaches \citep{cheng2025monster,xu2025igev++,wang2024selective,guo2025lightstereo,mons++}, we create the Basic Training Set (BTS) following MonSter \citep{cheng2025monster}, including Sceneflow \citep{sceneflow}, CREStereo \citep{li2022practical}, Tartan Air \citep{tartanair}, SintelStereo \citep{sintel}, FallingThings \citep{falling} and InStereo2K \citep{instereo2k}.
For supernet search set, we adopt 10\% pairs from Foundation Stereo Dataset (FSD) \citep{foundationstereo}.
We evaluate the KITTI2012 \citep{kitti2012}, KITTI2015 \citep{kitti2015}, Sceneflow \citep{sceneflow} and ETH3D \citep{eth3d} benchmarks.
we adopt the DrivingStereo \citep{yang2019drivingstereo} as zero-shot generalization validation following MonSter++ \citep{mons++}.

\begin{table}[t]\scriptsize
\setlength{\tabcolsep}{1.5pt}
  \centering
  \begin{tabular}{l|c|cccc|c|c}
    \toprule
    \textbf{Method}                          &\textbf{Iter. based} &\textbf{Sunny} & \textbf{Cloudy} & \textbf{Rainy} & \textbf{Foggy} & \textbf{Avg.} & \textbf{Latency(s)}\\
    \hline
    ZeroStereo\citep{zerostereo}             & \RX                  & 3.15  & 2.69 & 11.71 & 1.70     & 4.81 & OOM\\
    Nerf-Stereo\citep{nerfstereo}            & \RX                  & 2.88  & 2.95 & 8.47 & 3.41      & 4.43 & -\\
    SMoEStereo\citep{SMoEStereo}             & \RX                  & 3.51  & 3.11 & 6.08 & 4.77      & 4.37 & -\\
    IGEV\cite{xu2023iterative}               & \GC                 & 5.30 & 5.48 & 15.41 & 5.01       & 7.80 & 1.29\\
    Selective-IGEV\cite{wang2024selective}   & \GC                  & 5.67 & 5.35 & 13.84 & 4.79      & 7.41 & 1.61\\
    DefomStereo\citep{jiang2025defom}        & \GC                  & 3.39 & 3.13 & 12.84 & 2.57      & 5.48 & 5.05\\
    FoundaStereo\citep{foundationstereo}     & \GC                  & 3.22  & 2.81 & 11.20 & 2.50     & 4.93 & 14.02\\
    \rowcolor[rgb]{ 0.85, 1, 0.85}
    MonSter++\citep{mons++}                  & \GC                 & 2.60  & 2.12 & 3.08 & 2.94   & 2.69 & 7.63\\
    \hline
    HITNet(L)\cite{tankovich2021hitnet}      & \RX                 & 90.53 & 90.43 & 96.40 & 96.73    & 93.52 & 0.44\\
    IINet\cite{li2024iinet}                  & \RX                 & 17.73 & 24.26 & 38.10 & 30.69    & 27.70 & 0.29\\
    CoEx\cite{bangunharcana2021CoEx}         & \RX                 & 17.39 & 22.06 & 29.28 & 23.05    & 22.95 & 0.17\\
    LightStereo(S)\cite{guo2025lightstereo}  & \RX                 & 9.68  & 11.01 & 16.80 & 14.81    & 13.08 & 0.13\\
    RT-MonSter++\citep{mons++}               & \GC                  & 3.18 & 2.86 & 5.90  & 4.76      & 4.18 & 0.79\\
    RT-IGEV++\cite{xu2025igev++}             & \GC                  & 4.33 & 5.30 & 14.25 & 8.27      & 8.04 & 0.39\\
    \rowcolor[rgb]{0.85, 1, 0.85}
    \textbf{PipStereo (Ours)}                & \GC (Only 1)         & 3.27 & 2.69 & 7.71 & 3.76 & 4.35 &  0.44\\
    \bottomrule
  \end{tabular}
  \vspace{-5pt}
  \caption{
   \textbf{Zero-shot generalization performance.} Evaluation on DrivingStereo \cite{yang2019drivingstereo} under diverse weather conditions, reported using the D1-All metric.}
\vspace{-20pt}
  \label{tab:generalization}
\end{table}

\begin{figure}[t]
    \centering
    \includegraphics[width=0.49\textwidth]{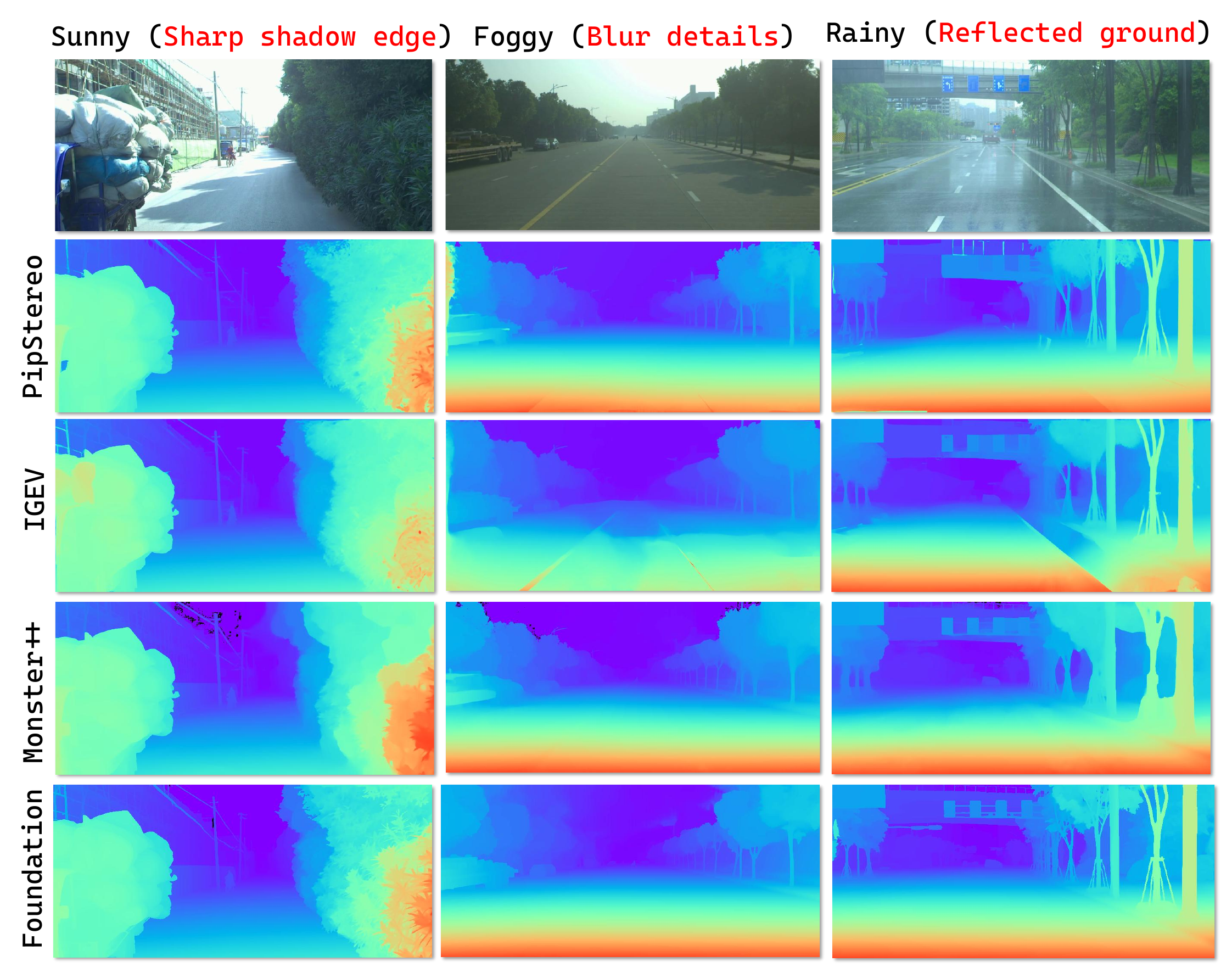}
    \caption{\textbf{Visual comparison.} Leveraging depth prior, PipStereo successfully handles ill-posed regions and produces sharper details and more coherent structures compared to IGEV\cite{xu2023iterative}.}
    \label{fig:visz}
    \vspace{-10pt}
\end{figure}

\noindent
\textbf{Implementation Details.}
We develop basic modules of PipStereo based on Selective-IGEV \citep{wang2024selective}. All models are trained and validated on a cluster of 8 NVIDIA RTX 4090 GPUs (24GB HBM). Whole model efficiency profiling is conducted on an NVIDIA Jetson Orin NX (16GB LPDDR5). FlashGRU is implemented using CUDA 11.4 within JetPack 5.3.
In Stage 1, we first warm up the supernet for 10K steps on the BTS, then freeze its weights and perform supernet search over 10K generations on 10\% of the FSD \citep{foundationstereo} using an early-exit strategy. Subsequently, we train for 200K steps to transfer monocular depth priors, with a learning rate of 2.5e-4, weight decay of 1e-5, and batch size 24. Task-specific finetuning on ETH3D \citep{eth3d} and KITTI \citep{kitti2012, kitti2015} benchmarks follows for 100K steps. In Stage 2, we apply the PIP algorithm and finetune only the update block for 50K steps with a learning rate of 2e-4 and batch size 64. For zero-shot generalization evaluation, both stages are trained on the full BTS as same as MonSter++\citep{mons++}. (Refer more details in Appendix).

\subsection{Integrated Performance}
\noindent
\textbf{In-Domain Comparison.} 
As illustrated in \cref{tab:benchmark}, we compare PipStereo with the high-accuracy iterative based approaches and real-time methods.
On SceneFlow \citep{sceneflow}, PipStereo outperforms the second-best method by a substantial margin of 13.5\%. On ETH3D \citep{eth3d}, it achieves significant gain: 73.4\% lower Bad-1 (Noc), 54.1\% lower Bad-1 (All), and 34.4\% lower RMSE (Noc).
PipStereo achieves accuracy on par with state-of-the-art methods that rely on multiple iterative refinements, yet delivers astonishing efficiency on edge platform, its inference is 22$\times$ faster than MonSter \citep{cheng2025monster}, 14$\times$ faster than DEFOM-Stereo \citep{jiang2025defom}, and a remarkable 41$\times$ faster than FoundationStereo-L \citep{foundationstereo}. This striking balance of precision and speed underscores its promise for resource-constrained edge deployment.

\noindent
\textbf{Zero-Shot Generalization.} 
\cref{tab:generalization} presents a comparison of zero-shot generalization performance across representative approaches. CoEx \cite{bangunharcana2021CoEx}, HITNet \cite{tankovich2021hitnet}, LightStereo \cite{guo2025lightstereo}, and IINET \cite{li2024iinet} are evaluated using models trained solely on SceneFlow; RT-MonSter++ \cite{mons++} and RT-IGEV++ \cite{xu2025igev++} leverage mixed-dataset pretraining; all other results are adopted from the reported numbers in \citet{mons++}. Notably, as evidenced by rows 1-4 of \cref{tab:generalization}, non-iterative real-time methods exhibit markedly limited generalization capability. In contrast, even lightweight iterative approaches demonstrate substantially stronger cross-domain robustness. This observation strongly suggests that iterative refinement plays an indispensable role in cultivating a model's capacity for generalization, acting not merely as a post-hoc refiner but as a core inductive bias during learning.

\begin{table}[t]\scriptsize
\setlength{\tabcolsep}{0.1pt}
\setlength{\arrayrulewidth}{0.6pt}
  \centering
  \begin{tabular}{c|c|c|c|c|c}
    \toprule
    \textbf{~ID~} & \textbf{Method} & \textbf{Alias} & \textbf{Infer iters.} & \textbf{SceneFlow\cite{sceneflow}} & \textbf{Latency(s)}\\
    \bottomrule
     \rowcolor[rgb]{ .886,  .937,  .851}
     1&Selective-IGEV\cite{wang2024selective}                 
                                               & -                    & 12 & 0.44 & 1.61 \\
     2&~~~-                                    & -                    & 32 & 0.44 & 3.57 \\
     3&~~~-                                    & -                    & 1  & 0.64 & 0.60 \\
     4&~~~+MPT (w./o. search)                  & -                    & 32 & 0.40 & 3.40 \\
     5&~~~+MPT                                 & PipStereo-i32        & 1  & 0.56 & 0.44 \\
     6&~~~+MPT                                 & PipStereo-i32        & 32 & 0.38 & 3.40 \\
     7&~~~+MPT, +PIP                           & PipStereo-p1         & 16 & 0.38 & 1.87 \\
     8&~~~+MPT, +PIP                           & PipStereo-p2         & 8  & 0.40 & 1.02 \\
     9&~~~+MPT, +PIP                           & PipStereo-p3         & 4  & 0.41 & 0.68 \\
     10&~~~+MPT, +PIP                          & PipStereo-p4         & 2  & 0.43 & 0.51 \\
     11&~~~+MPT, +PIP                          & PipStereo            & 1  & 0.45 & 0.44\\
     12&~~~+MPT, +PIP, +Flash                  & P2-Flash             & 8  & 0.43 & 0.76 \\
     13&~~~+MPT, +PIP, +Flash                  & P3-Flash             & 4  & 0.43 & 0.55 \\
     \bottomrule
     \rowcolor[rgb]{ .886,  .937,  .851}
     14&Raft-Stereo\citep{lipson2021raft}
                                                & -                   & 32 & 0.72 & 2.95\\
     15&~~~-                                    & -                   & 1  & 3.65 & 0.86 \\
     16&~~~-                                    & -                   & 4  & 2.22 & 1.07 \\
     17&~~~+PIP                                 & Raft-p5             & 1  & 2.16 & 0.86 \\
     18&~~~+PIP                                 & Raft-p3             & 4  & 1.43 & 1.07 \\
     19&~~~+PIP, +Flash                         & Raft-p3-Flash       & 4  & 1.47 & 0.69 \\
    \bottomrule
    \bottomrule
    \textbf{ID} & \textbf{Strategy} & \textbf{Alias} & \textbf{Infer iters.}  & \textbf{ETH3D\cite{eth3d}} & \textbf{Latency(s)}\\
    \bottomrule
    \rowcolor[rgb]{ .886,  .937,  .851}
    20&FoundaStereo\citep{foundationstereo}
                                                & -                   & 32 & 0.38 & 14.02\\
    21&~~~-                                     & -                   & 1  & 0.94 & 7.14 \\
    22&~~~+PIP                                  & Founda-p5           & 1  & 0.70 & 7.14 \\

    \bottomrule
  \end{tabular}
  \vspace{-5pt}
  \caption{
   \textbf{Integration ablations.}
MPT substantially boosts performance;
PIP effectively curbs accuracy degradation from fewer iterations,
while FlashGRU provides acceleration for methods that more than one iteration under PIP.
}
\vspace{-15pt}
  \label{tab:modabs}
\end{table}

\subsection{Ablation Study}
\cref{tab:modabs} compares 3 baselines (Selective-IGEV \cite{wang2024selective}, Raft-Stereo \cite{lipson2021raft}, and FoundaStereo \citep{foundationstereo}) and their variants on EPE of Sceneflow \cite{sceneflow} test set and Bad-1 (Noc) of ETH3D \cite{eth3d}. We use the aliases of variants in \cref{tab:modabs} for following content.

\noindent
\textbf{Effect of Monocular Prior Transfering (MPT).} 
Comparing rows 1-5, the monocular depth prior transfer learning paradigm yields an overall 13.6\% reduction in end-point error (EPE). The architecture obtained after supernetwork search demonstrates a notably improved convergence trend during pretraining and further achieves an additional 5\% EPE reduction (see the appendix for detailed configurations of the block sequence and training gain).
Comparing the Selective-IGEV baseline (with only a single inference iteration, row 3) against PipStereo-i32 (also using a single inference iteration, row 5), it is evident that the MPT provides a substantial ability to mitigate accuracy degradation.
We illustrate a zero-shot visual comparison of DrivingStereo\cite{yang2019drivingstereo} on different challenges in \cref{fig:visz}.

\noindent
\textbf{Effect of Progressive Iteration Pruning (PIP).} 
On row 5, we observe that directly applying PipStereo-i32 with a single inference iteration incurs an EPE increase of 0.18 (+32.1\%). In contrast, after finetuning with the PIP algorithm for 5 times, the EPE is increase to only 0.07 (+15.5\%), demonstrating that the PIP algorithm effectively mitigates accuracy loss caused by reducing the number of iterations.
Furthermore, we validate the effectiveness of PIP on both FoundationStereo \cite{foundationstereo}. As shown in rows 20-22, when FoundationStereo \cite{foundationstereo} is applied with only a single inference iteration, its Bad-1 error increases by 0.6 (+61.2\%). In contrast, after PIP fine-tuning (Founda-p5, row 22), the Bad-1 error rises by only 0.32 (+45.0\%), clearly demonstrating PIP's strong capability in suppressing accuracy degradation caused by reduced iteration counts.

\noindent
\textbf{Accuracy Degradation Study of PIP.} 
Comparing rows 5 through 11 further reveals the gradual accuracy degradation trend during the PIP finetuning process. Specifically, comparing rows 7 and 9 shows that compressing the iteration count from 32 to 16 has nearly negligible impact on accuracy, while further reduction leads to progressively increasing accuracy degradation. This observation aligns well with the hit-rate behavior of IGEV \cite{xu2023iterative} illustrated in \cref{fig:motivation}: beyond 10 iterations, the hit rate plateaus, and the degradation pattern of PipStereo-pk closely matches this empirical finding.

\noindent
\textbf{Speedup Compensation for Pruning Less Iterations.} 
The PIP algorithm demonstrates a clear ability to suppress accuracy degradation for methods that employ regularized initial disparity, e.g., FoundationStereo \cite{foundationstereo}, IGEV \cite{xu2023iterative}, and Selective-IGEV \cite{wang2024selective}. However, rows 14-17 reveal a limitation of PIP when applied to methods relying solely on initial disparity: despite some mitigation benefit, the EPE of Raft-p5 remains above 2.
As shown in rows 18, Raft-p3 retains a small number of iterations can effectively alleviate this accuracy drop, and by incorporating our proposed FlashGRU module, we achieve notable speedup: e.g., with only four iterations, Raft-p3-Flash enables a 1.55$\times$ acceleration over RAFT-Stereo \cite{lipson2021raft} with 4 iterations. Moreover, as evidenced in rows 11-13, FlashGRU is compatible with regularized initial disparity paradigm and consistently delivers approximately 1.23$\times$ inference speedup compared to conventional implementations at the same iteration count (e.g., P3-Flash and P2-Flash).

\begin{table}[t]\footnotesize
\setlength{\tabcolsep}{0.2pt}
  \centering
  \begin{tabular}{c|c|c|c|c}
    \toprule
    \textbf{Flash.} & \textbf{Size} & \textbf{RNN Latency} & \textbf{Mem. Peak (MiB)} & \textbf{Mem. Request}\\
    \hline
    \RX & \tiny\textbf{320$\times$736}   & 28.8ms                             & 743                         & \scriptsize2.260$\times e^6$\\
    \GC & \tiny\textbf{320$\times$736}   & \textbf{14.8ms}(\rbt{$\times$1.94}) & \textbf{561}(\gat{24.4\%})  & \scriptsize$\bm{0.947\times e^6}$(\gat{58.1\%})\\

    \RX & \tiny\textbf{640$\times$1472}  & 45.2ms                            & 1431                        & \scriptsize9.258$\times e^6$\\
    \GC & \tiny\textbf{640$\times$1472}  & \textbf{15.2ms}(\rbt{$\times$2.97}) & \textbf{715}(\gat{50.0\%})  & \scriptsize$\bm{2.189\times e^6}$(\gat{76.4\%})\\

    \RX & \tiny\textbf{1280$\times$2944} & 122.4ms                            & 4105                        & \scriptsize37.756$\times e^6$\\
    \GC & \tiny\textbf{1280$\times$2944} & \textbf{16.8ms}(\rbt{$\times$7.28}) & \textbf{957}(\gat{76.6\%})  & \scriptsize$\bm{7.224\times e^6}$(\gat{80.9\%})\\
    \bottomrule
    \end{tabular}
  \vspace{-5pt}
  \caption{
   \textbf{Efficiency gain of P3-Flash on NVIDIA RTX4090 24GB (Ada Lovelace).}
Baseline without "Flash" is the ConvGRU implementation of Selective-IGEV\cite{wang2024selective},
and tested on downsampled 4$\times$ feature map for update block.
Red marks the speedup ratio.
Green marks the reduction ratio.
The "Request" is the sum of "ld/st" micro requests to GPU global memory.
}
\vspace{-15pt}
  \label{tab:flash}
\end{table}

\subsection{Efficiency Gain of FlashGRU} 
In \cref{tab:flash}, we present efficiency metrics for FlashGRU (CUDA 12.1) on the NVIDIA RTX 4090, as reported by Nsight Compute 2024.07. Due to limitations with unified memory on NX Orin, detailed memory access profiling was not feasible. Using P3-Flash, we evaluate three input resolutions to assess peak GPU memory usage and inference speed at the iteration start and just before 4$\times$ upsampling.

\noindent
\textbf{Speedup of FlashGRU.}
We demonstrate the remarkable acceleration potential of FlashGRU on the Ada Lovelace architecture. As shown in the third column of the table, the speedup benefit of FlashGRU becomes increasingly pronounced as resolution scales up. At 2K resolution (i.e., 1280$\times$2944), FlashGRU achieves a latency speedup of up to 7.28$\times$ for RNNs. Even at the standard resolution of 320$\times$736, FlashGRU still delivers a 1.94$\times$ speedup.

\noindent
\textbf{Global Memory Access Analysis.}
Native RNNs suffer from frequent global memory reads and writes of hidden states, which introduce numerous fine-grained bubbles during a computation graph. This inefficiency is dramatically exacerbated at high-resolution feature maps: as shown in rows 1, 3, and 5 of \cref{tab:flash}, both peak memory consumption and global memory access requests surge sharply with increasing resolution. FlashGRU's performance gains stem primarily from aggressive memory-access optimization. As evidenced by rows 2, 4, and 6 of \cref{tab:flash}, in stark contrast, FlashGRU's memory-efficiency advantage scales favorably with resolution, becoming even more pronounced at higher resolutions. At 2K resolution (1280$\times$2944), FlashGRU reduces memory instruction dispatches by an order of magnitude, achieving a 76.6\% reduction in peak memory usage and an 80.9\% decrease in global memory accesses.

\section{Conclusion}
While iterative stereo matching delivers high accuracy, its dependence on RNNs poses a significant barrier to deployment on resource-constrained edge devices—a challenge largely overlooked in prior work. We observe that disparity updates during iterative refinement exhibit strong spatial sparsity and temporal redundancy. Leveraging this insight, we introduce three key innovations: (1) Progressive Iteration Pruning (PIP), which collapses recursive computation into an efficient near-single-pass inference; (2) a collaborative Monocular Prior Transfer (MPT) framework that implicitly integrates depth priors without requiring an additional monocular encoder; and (3) FlashGRU, a hardware-aware RNN operator co-designed with structured sparsity and memory efficiency. Together, these components enable real-time stereo matching that matches the accuracy of large iterative models while achieving exceptional efficiency on edge hardware—significantly outperforming existing real-time methods in both accuracy and cross-domain generalization.
{
    \small
    \bibliographystyle{ieeenat_fullname}
    \bibliography{main}
}


\end{document}